\documentclass[10pt,journal,compsoc]{IEEEtran}

\ifCLASSOPTIONcompsoc
  \usepackage[nocompress]{cite}
\else
  \usepackage{cite}
\fi

\usepackage{hyperref}       
\usepackage{url}            
\usepackage{booktabs}       
\usepackage{amsfonts}       
\usepackage{nicefrac}       
\usepackage{microtype}      
\usepackage{graphicx}
\usepackage{multirow}
\usepackage{enumitem}
\usepackage{subfig}

\usepackage{algorithm}
\usepackage{algpseudocode}
\usepackage{amsmath}
\usepackage{enumitem}
\usepackage{booktabs}
\usepackage{hyperref}
\usepackage{tabularx}

\usepackage{soul}
\soulregister\cite7 
\soulregister\ref7 

\begin{document}

\title{CSCAD: Correlation Structure-based Collective Anomaly Detection in Complex System}

\author{Huiling Qin, Xianyuan Zhan, Yu Zheng
\IEEEcompsocitemizethanks{
  \IEEEcompsocthanksitem Huiling Qin, Yu Zheng are with Xidian University and JD Intelligent Cities Research and JD Technology. E-mail: orekinana@gmail.com and msyuzheng@outlook.com \protect
  \IEEEcompsocthanksitem Xianyuan Zhan are with JD Intelligent Cities Research and JD Technology. E-mail: zhanxianyuan@gmail.com
}
}


\IEEEtitleabstractindextext{%
\begin{abstract}

	Detecting anomalies in large complex systems is a critical and challenging task. The difficulties arise from several aspects. First, collecting ground truth labels or prior knowledge for anomalies is hard in real-world systems, which often lead to limited or no anomaly labels in the dataset. Second, anomalies in large systems usually occur in a collective manner due to the underlying dependency structure among devices or sensors. Lastly, real-time anomaly detection for high-dimensional data requires efficient algorithms that are capable of handling different types of data (i.e. continuous and discrete). We propose a correlation structure-based collective anomaly detection (CSCAD) model for high-dimensional anomaly detection problem in large systems, which is also generalizable to semi-supervised or supervised settings. Our framework utilize graph convolutional network combining a variational autoencoder to jointly exploit the feature space correlation and reconstruction deficiency of samples to perform anomaly detection. We propose an extended mutual information (EMI) metric to mine the internal correlation structure among different data features, which enhances the data reconstruction capability of CSCAD. The reconstruction loss and latent standard deviation vector of a sample obtained from reconstruction network can be perceived as two natural anomalous degree measures. An anomaly discriminating network can then be trained using low anomalous degree samples as positive samples, and high anomalous degree samples as negative samples. Experimental results on five public datasets demonstrate that our approach consistently outperforms all the competing baselines.
\end{abstract}

}

\maketitle

\IEEEdisplaynontitleabstractindextext

\IEEEpeerreviewmaketitle

\ifCLASSOPTIONcompsoc
\IEEEraisesectionheading{\section{Introduction}\label{sec:introduction}}
\else
\section{Introduction}
\label{sec:introduction}
\fi
Efficiently detecting faults or anomalies is vital to safeguard the daily operation of many large complex systems. Typical applications including manufacturing system fault detection, network intrusion detection, abnormal bioactivities discovery and so on \cite{mukherjee1994network,abe2006outlier}. Although extensive anomaly detection studies have been conducted in the past, developing a robust anomaly detection technique for large systems with complex internal structures is still a challenging task to solve. The challenges arise from several aspects. 

First, unlike other data-driven tasks, collecting ground truth labels or prior knowledge for anomalies in complex systems is much harder. Anomalies are typically rare in the population, leading to highly unbalanced datasets. In many real-world systems, anomaly labels are collected through manual inspection, leading to very limited or even no anomaly labels being collected. In extreme cases, it is even impossible to know the basic information about the anomalies in the system, which forbids the use of many anomaly detection algorithms with a predetermined threshold (e.g. anomalous degree threshold or the proportion of anomalies in the data) \cite{candes2011robust,zhai2016deep,zong2018deep}.
Due to these facts, unsupervised anomaly detection algorithms \cite{tian2019learning} that can adaptively learn the discriminating boundaries of normal and anomaly samples are particularly desirable. 

\begin{figure}[thbp]
	\centering
	\includegraphics[width=3.35in]{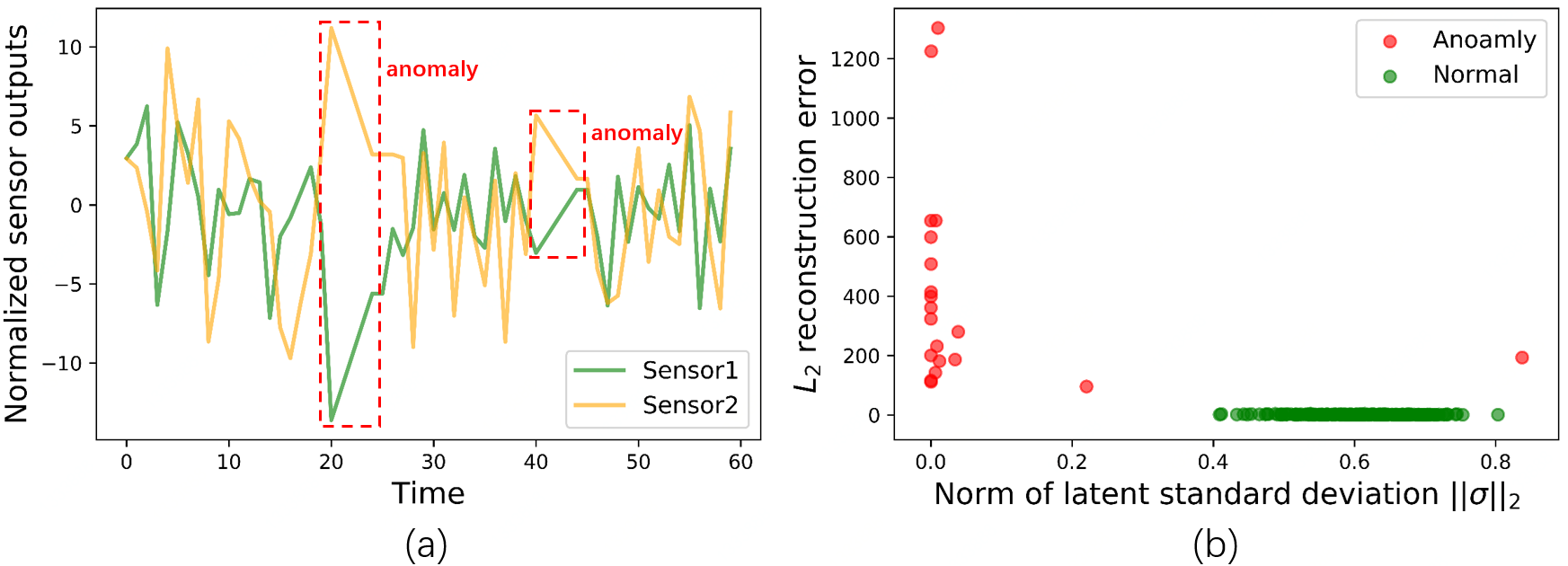}
	\caption{Sensory data and anomaly status from an human activity dataset: (a) Normalized readings of two sensors over time. (b) Reconstruction error and the norm of latent standard deviation vector of samples evaluated using a VAE.}
	\label{collective_anomaly_vis}
\end{figure}

Second, most sensors in large complex systems do not work independently. Many anomaly detection scenarios involve complicated internal dependency structures among sensors or devices. For example, sensors monitoring different parts of a manufacturing system typically have a complex interdependent relationship. Fault from a sensor can propagate to other dependent sensors, causing cascading failures in parts or the entire system. This is also referred to as \textit{collective anomaly} \cite{bontemps2016collective,araya2016collective,zheng2015detecting,zhang2018detecting}. 
Under such circumstances, an anomaly might not be that anomalous by checking a single data feature but could be identified as an anomaly when checking multiple related features simultaneously. Considering the feature space correlation to analyze the change in the underlying correlation structural pattern facilitates better modeling the operation characteristics of the system, leading to more accurate and robust results. This is particularly important in detecting early-stage anomalies when the anomalous pattern is not significant. 

Lastly, sensory data collected in large systems typically involve data with different properties (e.g. static and time-series) or types, such as continuous (e.g. temperature reading from a sensor) and discrete (e.g. on/off status of a switch) data. A unified approach is needed to handle different types of data and generalizable to time-series settings.

In this paper, we propose a highly adaptive correlation structure-based anomaly detection (CSCAD) framework to address the aforementioned challenges.
We consider the behavior of anomalies from three aspects: (1) break down of the original feature space correlation patterns (e.g. correlated features exhibit different correlation pattern); (2) anomalous samples are harder to reconstruct compared with normal samples, as they possess different statistical patterns; and (3) anomalous samples have larger variance when explaining using a model trained with the complete dataset. We illustrate the above anomalous behavior using a simple human activity dataset \cite{Dua:2019}. In Figure \ref{collective_anomaly_vis}(a), anomalies occur when two positively correlated sensors suddenly become uncorrelated; and in Figure \ref{collective_anomaly_vis}(b), the reconstruction error and latent standard deviation evaluated using a variational autoencoder (VAE) are observed to be highly discriminative features for the anomaly detection task.

Based on above observations, we use a graph convolutional variational autoencoder as the reconstruction network to capture the feature space correlation and reconstruction deficiency of samples, and further perform detection using a feedforward neural network as the discriminating network. Graph convolutional layers are used in the reconstruction network to mine the hidden structure in the feature space of data, which is constructed using a new correlation evaluation metric, the extended mutual information (EMI). It is capable of handling different types of data (continuous and categorical) and generalizable to time-series data. The feature space correlation provides extra information about the internal correlation structure among different features, thus enhances the data reconstruction capability of the model. Two anomalous degree measures, the reconstruction loss (measures reconstruction difficulty) and the latent standard deviation (measures internal variance) of samples can be obtained from the trained reconstruction network. The discriminating network is then trained using low anomalous degree samples as positive samples, and high anomalous degree samples or samples with anomaly labels as negative samples. This design allows the training process of the framework can be performed in an unsupervised manner without the need of introducing any predetermined anomaly threshold.

The contributions of our work is fivefold:

\begin{itemize}

	\item  We propose an extended version of mutual information to measure the correlation between data features with different characteristics. It is capable of handling different types of data (continuous and categorical) and generalizable to time-series data.
	
	\item We propose a novel graph-based deep learning framework, named CSCAD, to detect collective anomaly in complex systems. CSCAD leverages the correlation structure among features and the reconstruction deficiency of samples to perform anomaly detection.
	
	\item We use a discriminating network to adaptively identify anomalous samples by utilizing anomalous degree measures mined from the data in an unsupervised or semi-supervised manner, without the need of introducing any predetermined anomaly threshold.

	\item CSCAD provides a highly flexible framework, which is applicable to fully unsupervised, semi-supervised or supervised settings depending on the availability of anomaly labels, and can be easily extended to model high-dimensional time-series data.
	
	\item Experiments on five public datasets show that the proposed framework, as well as its time-series extension, consistently outperforms the baselines in terms of precision, recall and $F_1$ score, which demonstrate the effectiveness and extensibility of our framework.
	
\end{itemize}

\begin{figure*}[htbp]
	\centering
	\includegraphics[width=6.8in]{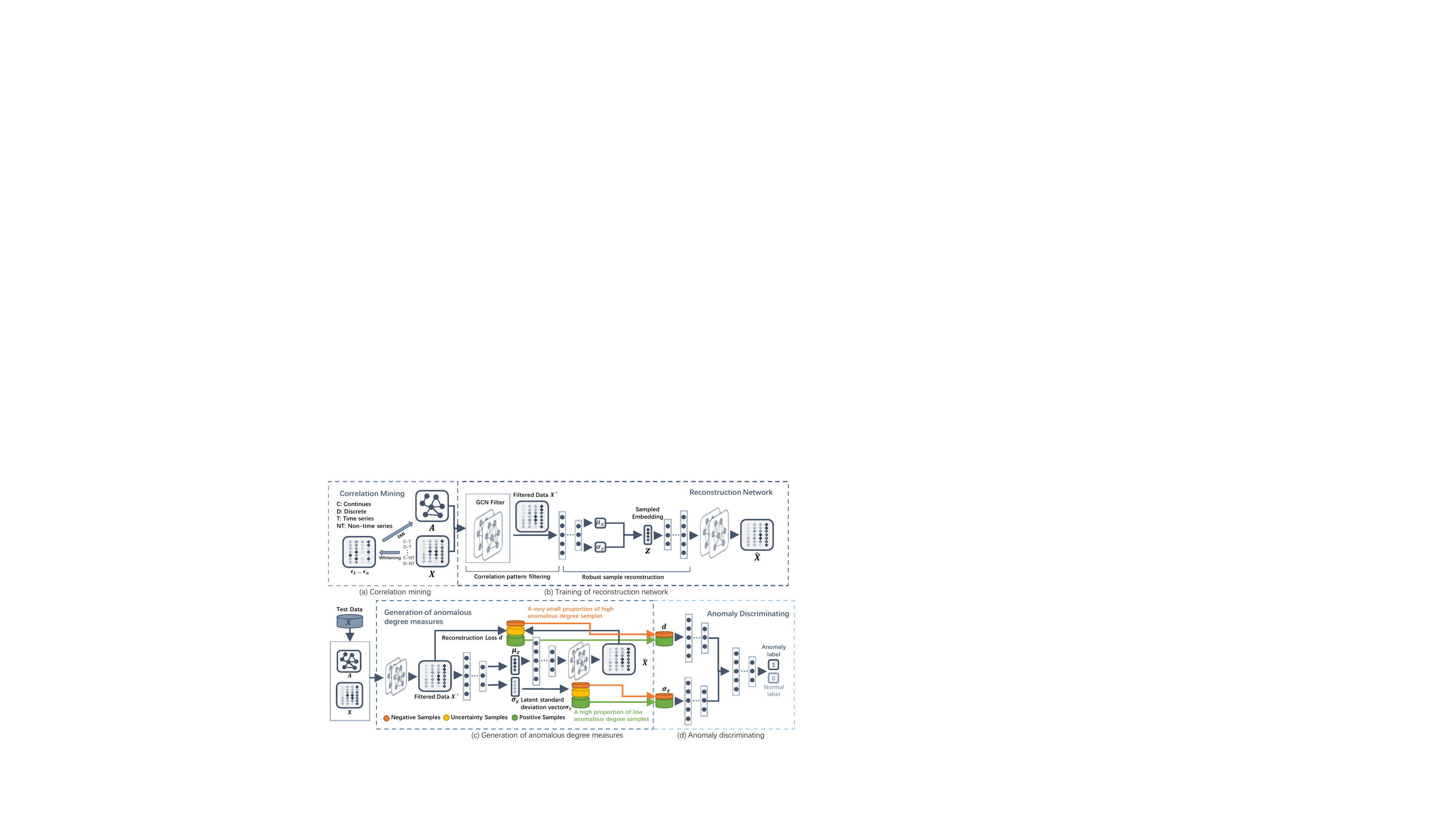}
	\caption{Proposed framework for collective anomaly detection}
	\label{framework}
\end{figure*}

\section{Overall Framework}\label{anomaly_detection_framework}

We consider the problem of collective anomaly detection in large complex systems. Let $\mathcal{X}$ be the space of all samples in a system, with the $i$th sample denoted as $X_i=(x_{i1},\cdots,x_{im})^T$, with $x_{ij}$ as the feature or sensor of sample $X_i$ in dimension $j$. 
We judge a sample $X_i$ to be normal or anomalous based on three criteria: (1) break down of usual feature space correlation pattern; (2) difficulty in reconstructing using a model trained using the complete dataset, and (3) latent space embedding exhibits high variance. 

Our framework adopts the following design to incorporate the three criteria. First, the hidden structure of the feature space is mined using a new extended mutual information (EMI) metric, which constructs a correlation structure graph among features. 
A reconstruction network modeled as graph convolutional variational autoencoder is then trained to generate two sample anomalous degree measures (reconstruction loss $d$ and latent standard deviation $\sigma_z$) by capturing the feature space correlation and perform robust sample reconstruction. 
The two anomaly degree measures are used as the inputs of a discriminative network to predict the final anomalous probability. The training of the discriminative network adopts a self-learning mechanism, which uses low anomalous degree samples as positive samples, and high anomalous degree samples or samples with anomaly labels as negative samples.
An illustration of the proposed framework can be found in Figure \ref{framework}.

\subsection{Correlation Structure Mining}

Large complex systems typically have large amount of sensors involving both continuous (e.g. temperature readings from a thermometer) and discrete (on/off status of a switch) data; some may be time varying and others are static. Most existing correlation measures, such as Pearson correlation coefficient, cross-variance and mutual information (MI), only work for data with identical types.
In order to mine the correlation structure across multiple types and properties of data features, a generic metric is needed.
We proposed the extended mutual information (EMI) metric extending the work of Galka et al. \cite{galka2006whitening} to measure the correlation between features with different characteristics. 

\begin{figure}[htbp]
	\centering
	\includegraphics[width=3.3in]{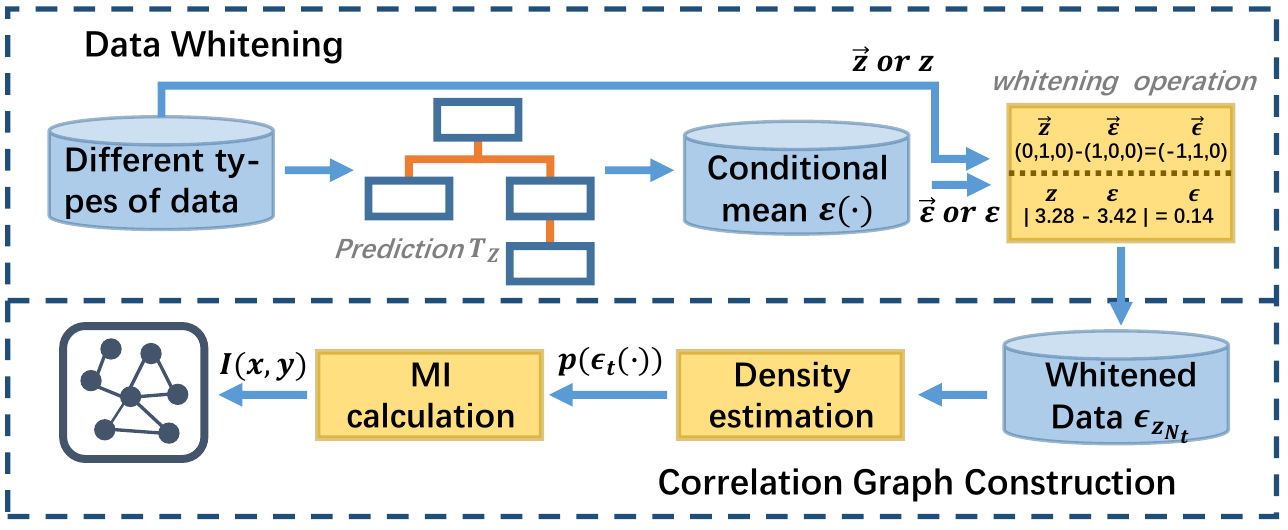}
	\caption{Illustration of the correlation mining process}
	\label{fig:emi}	
\end{figure}

EMI is based on information theory and directly works with probabilistic distributions of variables, which can be applied to both continuous and categorical data, and generalizable to time-series data. A llustration of the correlation mining process using EMI see Fig. \ref{fig:emi}

\textbf{Time-series MI}
The MI between two random variables $\mathbf{x}$ and $\mathbf{y}$ is defined as the Kullback-Leibler (KL) divergence between the joint distribution $p(\mathbf{x}, \mathbf{y})$ and the product of their marginals $p(\mathbf{x})$ and $p(\mathbf{y})$: $I(\mathbf{x}, \mathbf{y})=$$KL(p(\mathbf{x}, \mathbf{y})||p(\mathbf{x})p(\mathbf{y}))$. Galka et al. \cite{galka2006whitening} proposed an improved version of MI, which generalizes the MI for time-series data. Let $x_t$ and $y_t$ be two time-series random variables at time step $t$, the generalized MI can be written as:

\begin{equation}\label{mi}
	\begin{aligned}
		I(\mathbf{x}, \mathbf{y})=\log p((x_1,y_1),\cdots,(x_{N_t},y_{N_t}))-\\
		\log (p(x_1,\cdots,x_{N_t})p(y_1,\cdots,y_{N_t})
	\end{aligned}
\end{equation}

As the high-dimensional joint distributions $p((x_1,y_1), \cdots,$ $(x_{N_t},y_{N_t}))$, $p(x_1,\cdots,x_{N_t})$ and $p(y_1,\cdots,y_{N_t})$ are complicated to estimate, Galka et al. \cite{galka2006whitening} suggests removing the self-temporal correlation of the data by a whitening operation which utilizing a predictive model $\mathcal{E}(\cdot)$ to approximate the conditional means of variables, such that the whitened data satisfy the I.I.D. property (see \cite{galka2006whitening} for details). Let $\epsilon_{(\cdot)}$ represent the whitened form of data, which is the residual between the original and predictive conditional mean of data. 
The time-series MI can thus be evaluated as:

\begin{equation}\label{tmi}
	\begin{aligned}
		I(\mathbf{x}, \mathbf{y})=\log (p(\epsilon_{x_1},\epsilon_{y_1})\cdots p(\epsilon_{x_{N_t}},\epsilon_{y_{N_t}}))-\\\log (p(\epsilon_{x_1}) \cdots p(\epsilon_{x_{N_t}})p(\epsilon_{y_1}) \cdots p(\epsilon_{y_{N_t}}))
	\end{aligned}
\end{equation}

\textbf{Extended MI (EMI)}
With the nice statistical property of the whitened residuals, the key to compute the mutual information between two variables with different types lies in finding a unified predictive model to approximate the conditional means of the variable $\mathcal{E}(\cdot)$, as well as defining the residual for discrete variables. 

Based on time-series MI, we propose an extended version of MI to cover different types of data. There are two key ingredients of EMI: (1) a unified predictive model to perform temporal whitening for different types of data; and (2) a new scheme to properly define the ``residual'' of whitened discrete variables. 

For both continuous and discrete variables, we can perceive the conditional mean $\mathcal{E}(\cdot)$ as a temporal predictor $x_t=f(x_{t-1},x_{t-2},\cdots)$ that predict the data value at time $t$ given historical time-series information. 
In order to unify the predictive model for different types of data, we calculate the conditional mean using a regression tree (for continuous data) or a decision tree (for discrete data) given historical time-series data from a sliding time window. 
And for the conditional mean of the joint of two variables $\mathcal{E}((x_t,y_t)|(x_{t-1},y_{t-1}),(x_{t-2},y_{t-2}),\cdots)$, we can still use the regression tree regardless the types of $x_t$ and $y_t$, as the regression tree can handle both the continues and discrete data. 
The reasons that we choose the tree-based model to estimate the conditional means are because: (1) tree-based model can deal with different type of variable; (2) it can capture both the linear and non-linear relationship in the temporal data; (3) it is effective and simple, which help to reduce the computational cost during correlation evaluation for high-dimensional data.

Another challenge of evaluating mutual information between two variables across different data types is to properly define the residual for discrete variables, as simple subtraction is no longer well-defined for discrete variables. To address this issue while guarantee the linear transformation for the variable, we encode the discrete variables as one-hot encoded vector ($m$-dimensional vector if has $m$ discrete states), and calculate vector subtraction results $\vec y_i - \vec y_j$. There are a total of $m^2 - m + 1$ possible outcomes of $\vec y_i - \vec y_j$ (each element in the $m$-dimensional vector $\vec y_i - \vec y_j$ takes value of -1, 0, 1), therefore, we encode different outcomes of the vector subtraction as a new $m^2 - m + 1$ dimensional one-hot encoded vector $\vec z$ and use it as the residual for discrete variables. The residual is a lossless recording of the transition between different discrete states of the variables and is well-defined.

Finally, we can unify the whitening operation as follows:

\begin{equation}
	\epsilon_{z_{N_t}} = z_{N_t} \ominus T_{Z}(z_{N_t}|z_{N_t} \cdots z_{1})
\end{equation}

where $T_{Z}(\cdot)$ is the trained tree-based predictive model for variable $Z$; $\ominus$ is the subtraction operation when $z$ is continuous and is the previously defined discrete residual operation if $z$ is discrete.

After obtain the whitened residuals for the time-series data of the system, the marginal distribution of $p(\epsilon_t(x|x))$ and $p(\epsilon_t(x|(x,y)))$ can be estimated either by non-parametric statistics or methods such as kernel density estimation(KDE) (for continuous residuals) or simply count the proportion of each discrete residual state (for discrete residuals). For the joint distribution $p(\epsilon_t(x|x,y),\epsilon_t(y|x,y))$ involving both continuous and discrete variables, it is calculate as:

\begin{equation}\label{mix-jointdistribution}
\scalebox{0.98}{$
	p(\epsilon_t(x|x,y),\epsilon_t(y|x,y))=p(\epsilon_t(x|x,y)|\epsilon_t(y|x,y))p(\epsilon_t(y|x,y))
	$}
\end{equation}

where $x$ is a continuous variable, and $y$ is a discrete variable. The conditional probability of $p(\epsilon_t(x|x,y)|\epsilon_t(y|x,y))$ can be estimated by a set of non-parametric statistics conditioned on each discrete state of $y$. Finally, we can use Eq.\ref{mi}, \ref{tmi} to obtain the MI-based correlation for two time-series variables with arbitrary types.

With the extended data whitening scheme, the MI of data with different types can be evaluated in a similar procedure as discussed in \cite{galka2006whitening}. The proposed EMI allows evaluating correlation for both continuous and categorical data, and generalizable to time-series data. We use EMI to evaluate the pairwise correlation between different features of data, from which a correlation structure graph can be constructed, with nodes represent data features, and edges represent the pairwise correlation of the two features. We perform the pairwise EMI calculations in parallel to speed up the correlation graph construction.

\subsection{Reconstruction Network}
A notable characteristic of anomalies is that they are harder to reconstruct compared with normal samples \cite{zong2018deep}.
Inspired by this observation, we design a graph convolutional variational autoencoder as the reconstruction network for sample reconstruction, which is a combination of graph convolutional neural network (GCN) and variational autoencoder (VAE) \cite{Kingma2014AutoEncodingVB} (see Figure \ref{framework} (b)). We use GCN layers to capture the correlation structure of features mined using the EMI. Hence when data features exhibit distinct correlation patterns, the outputs filtered by the GCN layers will facilitate enlarging the reconstruction error evaluated by the VAE. Furthermore, we use the VAE rather than a conventional autoencoder.
As VAE can learn a latent mean and a latent standard deviation vector, which can be perceived as the denoised mean and the internal variance of the sample embedding, thus extracts more detailed information about the anomalous behavior of a sample. 

We use the GCN layer proposed in Defferrard et al. \cite{defferrard2016convolutional}
to model the correlation structure in the feature space. Consider a spectral convolution on graph defined as the multiplication of a graph signal $X \in \mathbb{R}^{m\times c}$ with a filter $g_\theta $ parameterized by $\theta $ in the Fourier domain:

\begin{equation}
	g_\theta \star _{\mathcal{G}} X=g_\theta (L)X=g_\theta (U\Lambda U^T)X=U g_\theta (\Lambda)U^T X
\end{equation}

where $U \in \mathbb{R}^{m\times m}$ is the matrix of eigenvectors, and $\Lambda \in \mathbb{R}^{m \times m}$ is the diagonal matrix of eigenvalues of the normalized graph Laplacian $L=I_N-D^{\frac{1}{2}}\Lambda D^{\frac{1}{2}}=U\Lambda U^T$, where $I_N$ is the identity matrix, $D \in \mathbb{R}^{m \times m}$ is the diagonal degree matrix with $D_{ii}=\sum_j A_{ij}$ and $A$ is the adjacency matrix of the correlation structure graph. A non-parametric filter, i.e. a filter whose parameters are all free, would be defined as:

\begin{equation}
	g_\theta (\Lambda) =diag(\theta)
\end{equation}

where the parameter $\theta \in \mathbb{R}^{n}$ is a vector of Fourier coefficients.

Directly perform convolution operation using above formulation is computationally expensive. Defferrard et al. \cite{defferrard2016convolutional} used the $K$th-order polynomial in the Laplacian to enable fast evaluation, which restricts the GCN to capture the information at maximum K step away from the central node (K-localized). The corresponding graph convolutional operator
and the $l$th layer output of GCN $H^{(l)} \in \mathbb{R}^{m\times d}$ given activation function $f(\cdot)$ are given as:

\begin{equation}
	g_\theta(L) X = \sum_{k=0}^{K-1}\theta_kL^k X,\quad H^{(l+1 )}=f\Big(\sum_{k=0}^{K-1}\theta_kL^k H^{(l)}\Big)
\end{equation}

The subsequent VAE component of the reconstruction network takes the outputs from the GCN layer to perform a robust reconstruction. VAE forces its encoder to generate a latent vector $z$ that roughly follow a Gaussian distribution, which is parameterized by a latent mean vector $\mu_z$, and a latent standard deviation vector $\sigma_z$. $\mu_z$ and $\sigma_z$ can be perceived as embeddings of the denoised mean and the internal uncertainty level of a sample,
which provides important information about the anomalous degree of the sample. 
Given a dataset of $N$ samples, the reconstruction network can be trained in a similar manner as VAE using variational inference \cite{Kingma2014AutoEncodingVB} by minimizing the following objective function:

\begin{equation}
	\begin{aligned}
		J(\theta, \phi)=\frac{1}{N}\sum^N_{i=1}L(X_i,\hat{X}_i)+\frac{\lambda}{N}\sum^N_{i=1}D_{KL}(q_{\phi}(z_i|X_i)||p_\theta (z_i))
		\end{aligned}
\end{equation}

\begin{figure*}[htbp]
	\centering
	\includegraphics[width=7.2in]{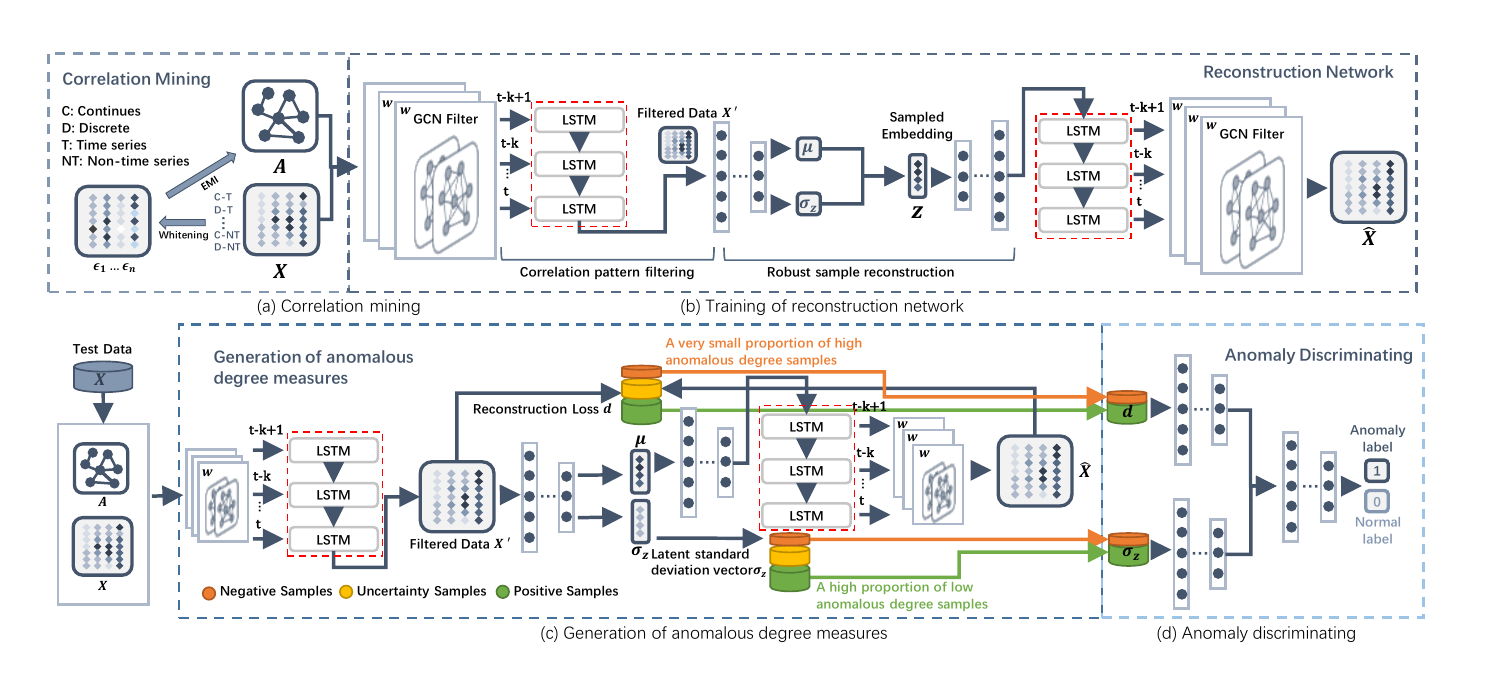}
	\caption{Proposed framework for collective anomaly detection in time-series settings}
	\label{fig:framework-2}
\end{figure*}

where $L(X_i,\hat{X}_i)=||X_i-\hat{X}_i||^2_2$ is the loss between original sample $X_i$ and reconstructed sample $\hat{X}_i$. 
The KL divergence $D_{KL}(q_{\phi}(z|X)||p_\theta (z))$ forces the approximated posterior distribution $q_{\phi}(z|X)$ to be similar to the prior distribution $p_\theta (z)$, which improves the robustness of reconstruction when separating the internal noise from input data. 

The proposed reconstruction network can be easily extended to model high-dimensional time-series data. This is done by introducing the recurrent neural network layer (e.g. LSTM, GRU) after GCN layer to capture temporal information across different time steps. 


\subsection{Discriminating Network and Detection Strategy}
\textbf{Generation of anoamlous degree measures.} Utilizing the trained reconstruction network, we derive two anomalous degree measures to support anomaly detection:
\begin{itemize}[leftmargin=*]
	\item The reconstruction loss $\vec{d_i}=\big[(x_{i1}-\hat{x}_{i1})^2,\cdots,(x_{im}-\hat{x}_{im})^2\big]^T$ is 
	the element-wise reconstruction loss calculated by original sample $X_i$ and reconstructed sample $\hat{X}_{i}$.
	We generate $\hat{X}_{i}$ by applying the decoder of GCVAE on the latent mean vector $\mu_z$ rather than the sampled latent vector $z$. As $\hat{X}_{i}$ generated from $\mu_z$ is deterministic and can be perceived as a denoised version of $X_i$, which helps reconstruction loss $d$ more accurately reflect the reconstruction difficulty of a sample. As a result, a sample with a larger $d$ indicates a higher anomalous degree.
	\item The latent standard deviation $\sigma_z$ represents the internal variance level of a sample. It is another anomalous degree measure which reflects the uncertainty of the input data. A sample with high uncertainty or significantly deviate from the regular patterns of the data is likely to be an anomaly.
\end{itemize}
These two anomalous degree measures provide more detailed anomalous information of a sample to allow users to “interpret” the results. The reconstruction loss and latent standard deviation are given as a vector indicating the reconstruction difficulty and uncertainty level of each feature (e.g. each sensor), which can also help locate the most problematic sensors in the system. 

\textbf{Anomaly discriminating.} We construct a discriminating network using reconstruction loss $d$ and latent standard deviation $\sigma_z$ as inputs (see Figure \ref{framework} (c), (d)). In the discriminating network, $d$ and $\sigma_z$ are first fed into two sets of fully connected layers. Their outputs are then concatenated and fed into two fully connected layers to output the final anomalous probability of a sample. 

Training the discriminating network requires anomaly labels in the data, which can be difficult to acquire. We take an alternative approach by utilizing the already obtained anomalous degree measures $d$ and $\sigma_z$. We first evaluate $d$ and $\sigma_z$ of all samples in the training set, and rank the samples according to their norms ($||d||_2, ||\sigma_z||_2$). 
Considering that the anomalies are typically rare in the population, and a sample with large $d$ or $\sigma_z$ might be anomalous. We label 50\% (the proportion can be flexibly adjusted according to different real-world datasets) of low anomalous degree samples as positive samples. The negative samples are selected based on a very conservative criterion, that we only select a very small proportion of high anomalous degree sample (e.g. top 2.5\%)  as negative samples. The discriminating network is trained only using selected positive and negative samples.
When parts or full anomaly labels are available, the actual anomalies can be used as negative samples, which enables the proposed framework adaptable to semi-supervised or supervised settings. The proposed strategy does not need to specify a predetermined anomaly threshold that typically used in anomaly score-based methods \cite{zhai2016deep,zong2018deep,an2015variational,xu2018unsupervised,ikeda2018estimation}, and capable of learning complex high-dimensional separation boundaries between normal and anomalous samples.

At the detection stage, the reconstruction loss $d$ and latent standard deviation vector $\sigma_z$ are obtained using the trained reconstruction network. They are fed into the trained discriminating network to evaluate the final anomalous probability of the given sample (see Figure \ref{framework} (d) for detailed illustration).

\subsection{Time-Series Extension}
The proposed CSCAD framework can also be generalized to time-series settings. 
We focus on the collective anomaly detection problem that finds anomalies at time step $t+1$ based on previous $k$ time steps' data ($t-k+1, \cdots, t$). Let the sample at time step $t$ be $X_{t}=(x_{t1},x_{t2},\cdots,x_{tm})^T$. 
We use EMI to construct the feature space correlation structure graph of the time-series data. EMI can capture the correlation between features while eliminating the historical influence of the sequence. The reconstruction network is modified to include an LSTM layer to capture the temporal characteristics of the data.


The time-series data $X_t$ at each time step is first fed into the GCN layer, and the sequence of the GCN outputs are then modeled using the LSTM layer. The VAE component takes the final output of the LSTM layer and generates the latent mean and standard deviation of samples to support anomaly detection. The discriminating network and detection strategy remain the same as in the previous section. A illustration of the time-series extension framework see Fig. \ref{fig:framework-2}

\section{Experimental Result} 
In this section, we use public benchmark datasets to demonstrate the effectiveness of the proposed CSCAD framework against several competing baselines. 

\subsection{Dataset}
We use five public datasets 
from UCI machine learning repository \cite{Dua:2019} to evaluate the proposed framework, including KDDCUP, Thyroid, MoCap, UJIIndoorLoc for the static version of the framework, and Heterogeneity for the time-series extension of the framework. 
Detailed statistics of each dataset is presented in Table \ref{tab:graph_sample}.

\begin{table}[!th]
	
	\centering
	\caption{Statistics of the four public datasets}
	\label{tab:graph_sample}
	\begin{tabular}{cccc}
		\toprule
		& Dimensions & Instances & Anomaly ratio \\ 
		\midrule 
		KDDCUP99 & 121 & 494,021 &	20\% \\  
		Thyroid & 13 & 19,016 &	10\% \\ 
		MoCap & 36 & 15,963 &	8.5\% \\ 
		UJIIndoorLoc & 522 & 10,000 & 7.5\% \\
		Heterogeneity & 6 & 54,000 &	20\% \\ 
		\bottomrule
	\end{tabular} 
\end{table}

\begin{table*}[!tbh]
	\centering
	\caption{Experiment results of our framework and the baseline methods for four static datasets
	}
	\label{tab:accuracy}
	\scalebox{0.99}{
		\begin{tabular}{l|ccc|ccc|ccc|ccc}
			\toprule
			\multirow{2}{*}{Methods} & \multicolumn{3}{c|}{KDDCUP} & \multicolumn{3}{c|}{Thyroid} & \multicolumn{3}{c}{MoCap} & \multicolumn{3}{c}{UJIIndoorLoc } \\ 
			& Precision & Recall & $F_1$ & Precision & Recall & $F_1$ & Precision & Recall & $F_1$ & Precision & Recall & $F_1$ \\ 
			\midrule
			DAGMM & 0.830 & 0.839 & 0.835 & 0.273 & 0.257 & 0.265 & 0.431 & \textbf{1} & 0.603 & 0.720 & 0.709 & 0.714 \\ 
			AE-LOF & 0.456 & 0.461 & 0.458 & 0.269 & 0.507 & 0.351 & 0.122 & 0.283 & 0.171 & 0.067 & 0.264 & 0.107 \\ 
			LOF & - & - & - & 0.248 & 0.467 & 0.324 & 0.123 & 0.286 & 0.172 & 0.134 & 0.528 & 0.214 \\ 
			IF & 0.449 & 0.454 & 0.451 & 0.280 & 0.526 & 0.365 & 0 & 0 & 0 & 0.508 & 0.998 & 0.673 \\ 
			OC-SVM & - & - & - & 0.331 & 0.340 & 0.335 & 0.001 & 0.002 & 0.001 & - & - & - \\ 
			\midrule
			VAE-DN & 0.852 & 0.954 & 0.9 & 0.268 & 0.345 & 0.302 & 0.256 & 0.522 & 0.344 & 0.112 & 0.701 & 0.211 \\ 
			CSCAD(no $\sigma$) & 0.734 & \textbf{1} & 0.846 & 0.440 & 0.586 & 0.503 & \textbf{0.557} & 0.684 & 0.614 & 0.059 & \textbf{1} & 0.112 \\
			CSCAD(7.5\%) & 0.857 & 0.994 & 0.920 & \textbf{0.496} & \textbf{0.795} & \textbf{0.611} & 0.486 & \textbf{0.991} & 0.652 & \textbf{0.924} & 0.856 & \textbf{0.889} \\
			CSCAD(5\%) & 0.862 & 0.921 & 0.890 & 0.480 & 0.662 & 0.557 & 0.496 & 0.956 & \textbf{0.653} & 0.916 & 0.858 & 0.886 \\
			CSCAD(2.5\%) & \textbf{0.881} & 0.996 & \textbf{0.934} & 0.495 & 0.553 &0.522 & 0.507 & 0.792 & 0.618 & 0.706 & 0.894 & 0.789 \\ 
			\bottomrule
		\end{tabular} 
	}
\end{table*}

\begin{itemize}[leftmargin=*]
	\item \textbf{KDDCUP}. 
	The KDDCUP99 dataset consists of 34 continuous and 7 categorical features. The categorical features are encoded using one-hot encoding, resulting in a dataset of 121 dimensions. As only 20\% of samples are labeled as “normal” and 80\% are labeled as “attack”, therefore, we treat “normal” samples as anomalies in this task.
	
	\item \textbf{Thyroid}. 
	The Thyroid dataset is a disease dataset in which "negative" samples are treated as normal and others are anomalies. 7 continuous and 2 categorical relevant features are used in this task. Again, the categorical features are encoded using one-hot encoding.

	\item \textbf{MoCap}. 	
	The MoCap dataset consists of 36 continuous features. Representing the hand postures by 12 measuring points in the three-dimensional space. Here, we consider the hand posture "5" as anomalies and "1" as normal samples.

	\item \textbf{UJIIndoorLoc}. The UJIIndoorLoc is a multi-building indoor localization dataset containing 522 WiFi fingerprints as attributes. We consider "BUILDING ID" labeled "2" as normal samples and "0" as anomaly.
	
	\item \textbf{Heterogeneity}. 
	The Heterogeneity is a human activity recognition dataset which contains records of gyro sensor and accelerometer from smartphones and smartwatches, including 6 continuous time-series features. We consider "Stand" activities as normal samples, and "Stair Up" activities as anomalies.

\end{itemize}

\subsection{Baseline}
We consider several state-of-the-art deep learning approaches and a few widely used anomaly detection methods as baselines. The variants of CSCAD is also evaluated to support the ablation study of the proposed framework.

\begin{itemize}[leftmargin=*]
	
	\item \textbf{DAGMM}. The deep autoencoding Gaussian mixture model (DAGMM) \cite{zong2018deep} uses latent low-dimensional representation and a Gaussian mixture model to perform density estimation of data, and further predict anomalies using a predetermined threshold. It is a state-of-the-art method in the high-dimensional anomaly detection problem. As we use the same dataset with DAGMM, we remain the configurations in it's paper.
	
	\item \textbf{LOF}. The local outlier factor (LOF) \cite{breunig2000lof} finds anomalous data points by measuring the local deviation of a given data point with respect to its neighbors.
	
	\item \textbf{AE-LOF}. This method adopts a two-step approach. It first trains a deep autoencoder to generate a latent representation of a sample, then uses LOF to detect the anomaly.
	
	\item \textbf{IF}. Isolation Forest (IF) \cite{liu2008isolation} is a decision tree-based ensemble method. It partitions the sample points by randomly selecting a split value between the maximum and minimum values of a feature. Anomalies are more easily separated under random splits. This baseline is also used in the experiment on time-series data.
	
	\item \textbf{OC-SVM}. One-class support vector machine (OC-SVM) \cite{chen2001one} is a kernel-based method which learns a decision function between normal and anomalous samples.
	
	\item \textbf{VAE-DN}. This is a variant of the proposed CSCAD framework, which uses VAE as the reconstruction network without considering the feature space correlation.
	
	\item \textbf{CSCAD(no $\sigma$)}. This variant of the proposed framework only uses the reconstruction loss $d$ in the discriminating network, without considering the latent standard deviation vector $\sigma_z$.
	
	\item \textbf{CSCAD($p$\%)}. These models are the proposed framework of which the discriminating network is trained by selecting different percentages (i.e. 2.5\%, 5\%, 7.5\%, smaller than the actual anomaly ratio) of high anomalous degree samples as negative samples. 
	
	\item  \textbf{AE(LSTM)-IF}. AE(LSTM)-IF introduces LSTM layer in the autoencoder, and uses IF to perform detection on the encoded representation.
	
	\item  \textbf{VAE(LSTM)-DN}. VAE(LSTM)-DN is a variant of the time-series extension of CSCAD, which removes GCN layers in the reconstruction network.
	
	\item \textbf{SR-CNN}. SR-CNN borrows the SR model from visual saliency detection domain to time-series anomaly detection and then uses CNN as a discriminating network to improve the performance of the SR model. It is the state-of-the-art method in time-series anomaly detection problems. In the experiment, we use the different channel to capture the pattern of different features.

\end{itemize}

For the anomaly score-based (with predetermined threshold) methods, the threshold we choose for KDDCUP99 is $0.2$ and for Thyroid, MoCap, UJIIndoorLoc is $0.1$ which is close to the true anomaly rate of the dataset. Other parameters used in these algorithms remain the default. And network architecture of the deep learning-based approaches and the variants of CSCAD is the same as CSCAD. The model configurations of the proposed CSCAD is introduced as the next part.

\subsection{Model Configurations}

The detailed model configurations used on each individual datasets are summarized below. Let $m$ be the dimension of data features.

\begin{itemize}[leftmargin=*]
	
	\item \textbf{Reconstruction Network}. For all datasets, the reconstruction network runs with GCN($m$, k=2, ReLU)-FC($m$, 60, ReLU)-FC(60, 30, ReLU)-FC(30, 10, ReLU)-2 FC(10, 5, none)-reparameterize-FC(5, 10, ReLU)-FC(10, 30, ReLU)-FC(30, 60, ReLU)-FC(60, $m$, ReLU)-GCN($m$, k=2, none).
	
	\item \textbf{Discriminating Network}. The discriminating network contains 2 compression sub-networks that encode $d$ and $\sigma_z$.
	The outputs of the two compression sub-networks are concatenated and fed into a sub-discriminating network to output the final anomalous probability of a sample. For all datasets, the compression sub-network that encodes the $d$ runs with BN($m$, elu)-FC($m$, $m$/2, elu)-FC($m$/2, $m$/4, elu)-FC($m$/4, 10, elu)-FC(10, 5, elu); and the compression sub-network for $\sigma_z$ runs with BN($m$, elu)-FC($m$, $m$/2, elu)-FC($m$/2, 2, elu). Finally, the sub-discriminating network runs with BN(7, elu)-FC(7, 4, elu)-FC(4, 2, softmax).

	\item \textbf{Time-series extension framework}. We add an LSTM layer in the reconstruction network runs with LSTM($m$, layers=2, ReLU), other configurations are the same with CSCAD.

\end{itemize}

\subsection{Evaluation Metrics}
We use the precision, recall and $F_1$ score to evaluate the performance of the proposed CSCAD framework as well as the baselines, which are calculated as follows:

\[
Precision = \frac{TP}{TP+FP}
\]

\[
	Recall = \frac{TP}{TP+FN}
\]

\[
	F_1 = \frac{2*Precision*Recall}{Precision+Recall}
\]
where true positive (TP) is the number of anomalous samples that are correctly identified; false negative (FN) is the number of normal samples that are wrongly identified as anomalies; and false positive (FP) is the number of anomalous samples that are wrongly identified as normal samples. The higher values of $F_1$ score, recall, and precision correspond to more accurate anomaly detection.

\begin{figure*}[tbp]
	\centering
	\subfloat[KDDCUP99]{
		\includegraphics[width=1.69in]{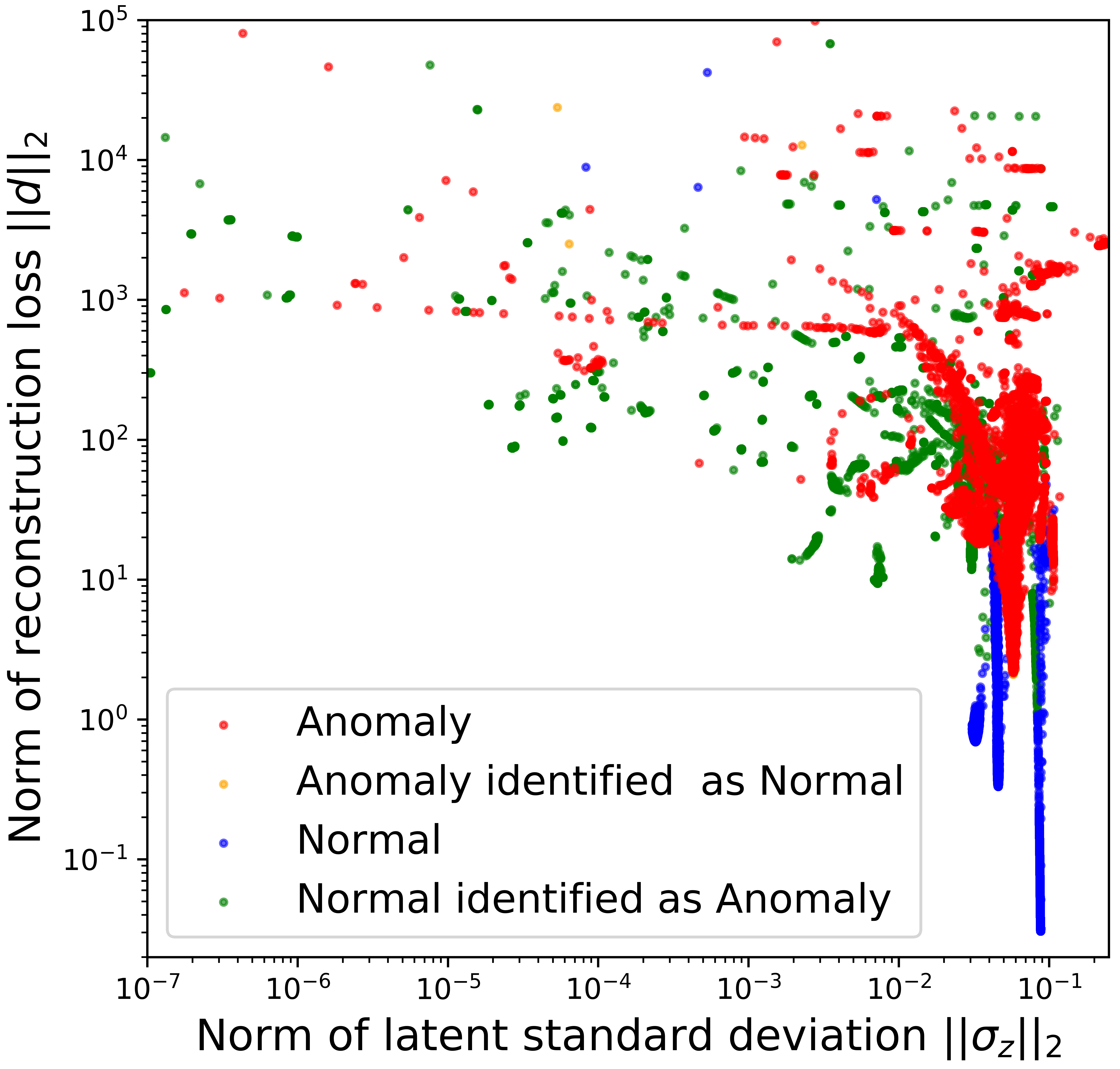}
	}
	\subfloat[MoCap]{
		\includegraphics[width=1.69in]{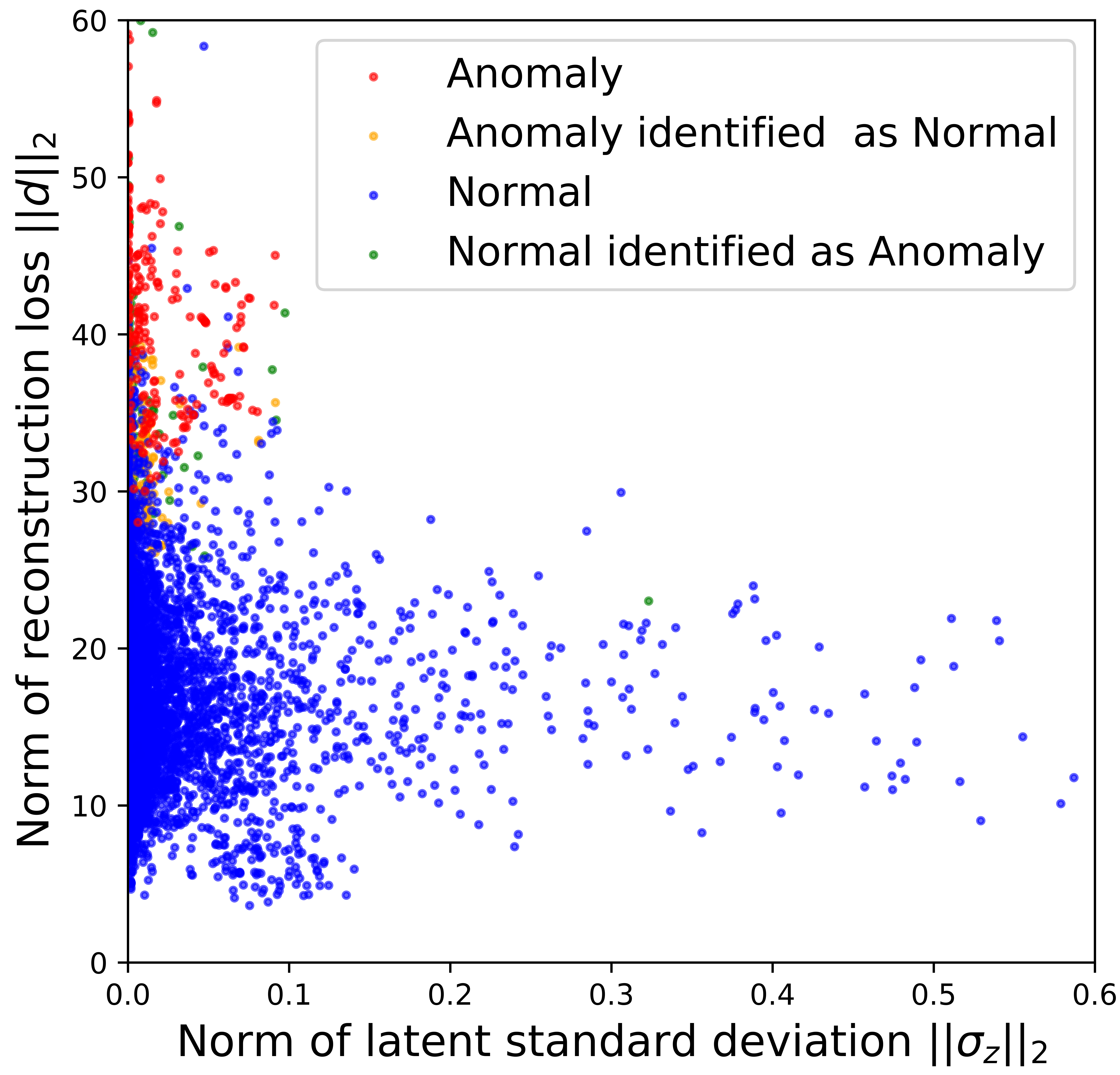}
	}
	\subfloat[Thyroid]{
		\includegraphics[width=1.67in]{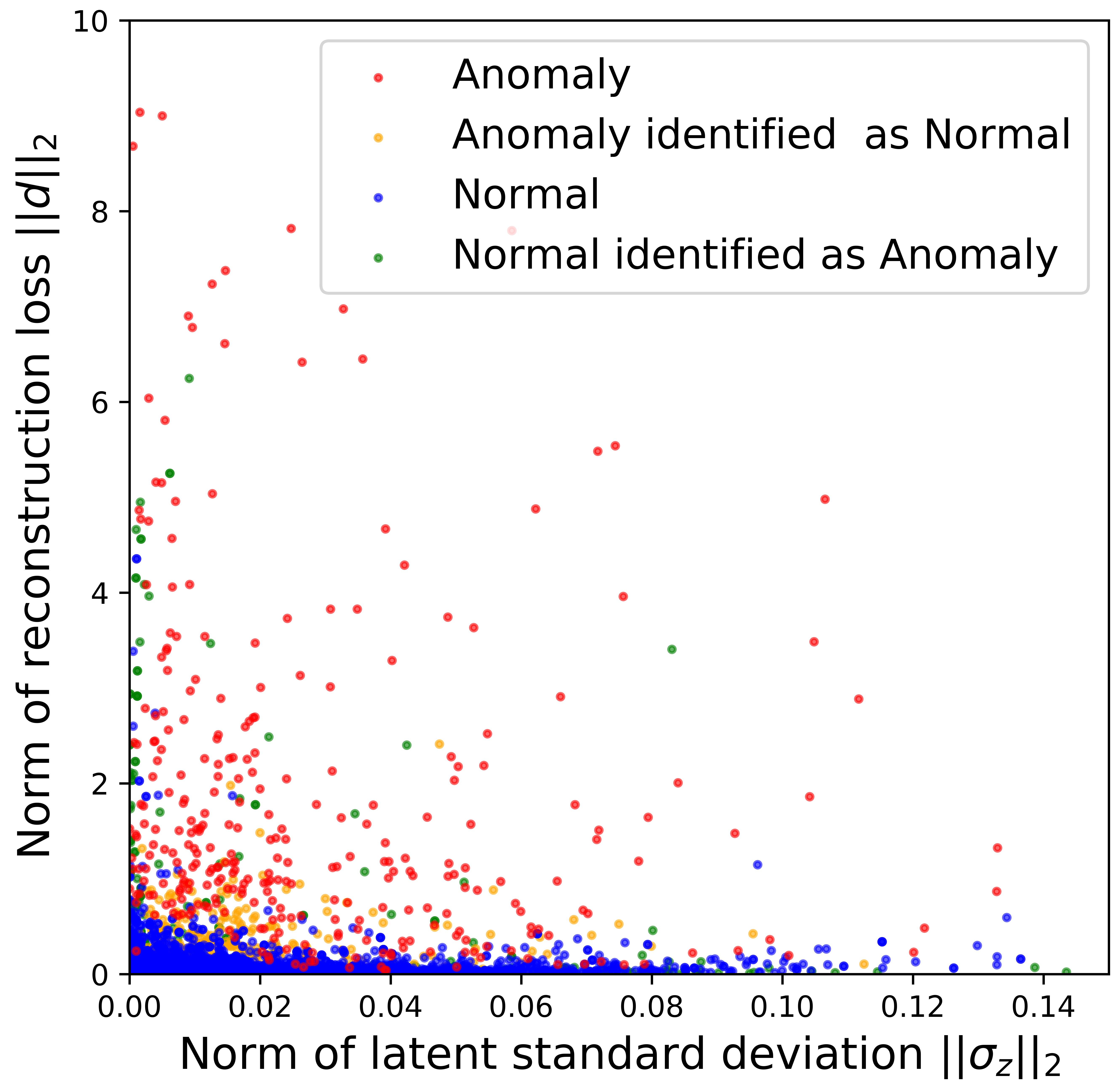}
	}
	\subfloat[UJIndoorLoc]{
		\includegraphics[width=1.75in]{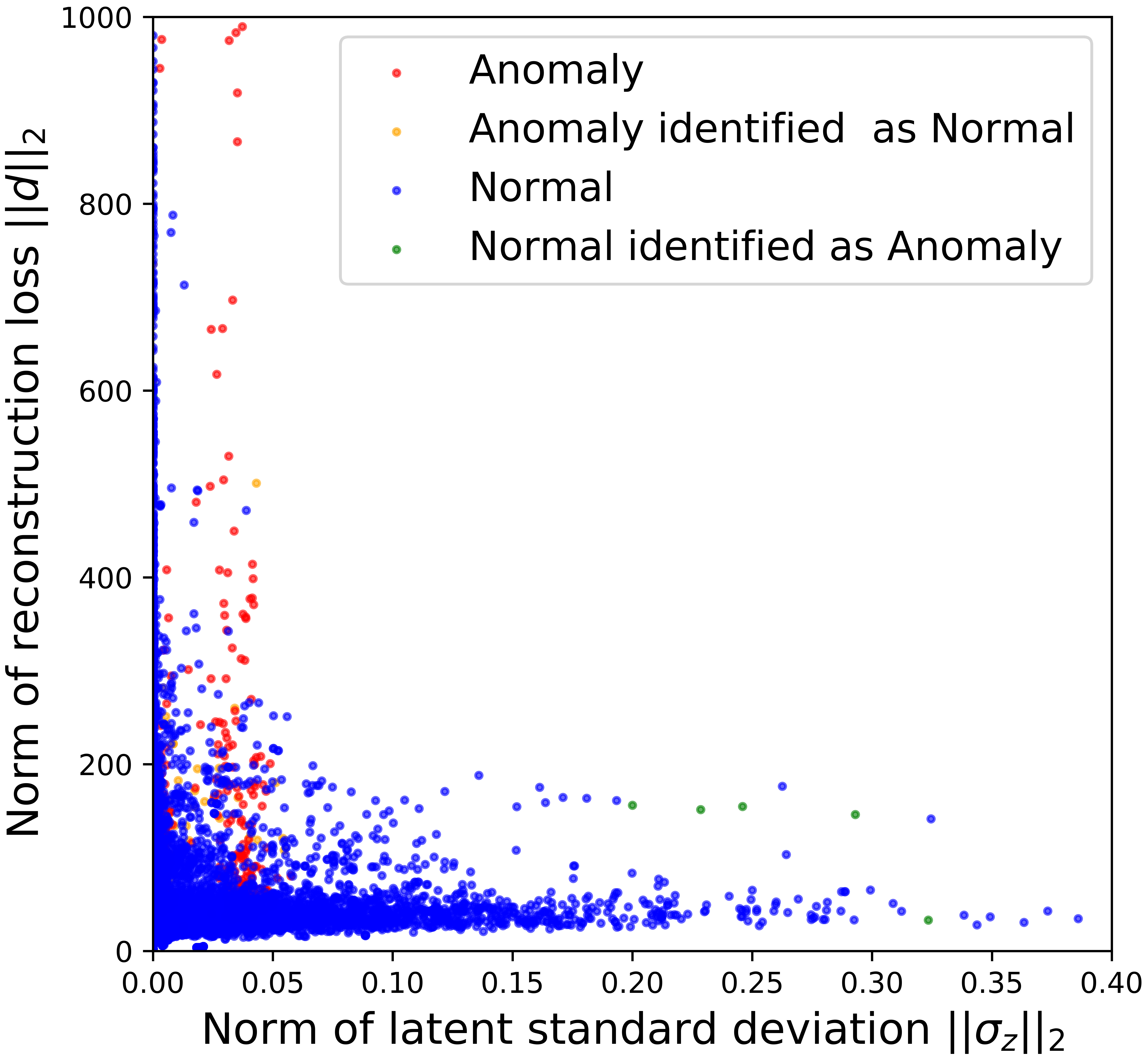}
	}
	\caption{Visualization of the detection results with respect to norms of reconstruction loss $d$ and latent standard deviation $\sigma_z$}
	\label{dataset_vis}
\end{figure*}

\subsection{Result}

\subsubsection{Detection accrancy} Table \ref{tab:accuracy} presents the precision, recall and $F_1$ score of the baseline methods and our proposed framework for different datasets.
It is shown that CSCAD demonstrates superior performance over all the baseline methods. 
Several baseline methods such as LOF and OC-SVM even failed within an acceptable running time due to the large data size (e.g., KDDCUP, UJIndoorLoc).
It is observed that even training by selecting only the most conservative 2.5\% high anomalous degree samples as negative samples, the proposed framework (CSCAD($2.5$\%)) achieves 10.2\%, 15.7\%, 1.5\% and 7.5\% improvement on $F_1$ score on different datasets over the best baseline methods (excluding VAE+DN and other CSCAD variant models). Moreover, comparing the results of CSCAD($2.5$\%, $5$\% and $7.5$\%), it is observed that conservatively selecting a very small proportion (much smaller than the anomaly ratio of the dataset) of high anomalous degree samples as negative samples for training still enables the discriminating network to maintain reasonable anomaly detection accuracy.
This is important, as in many real-world scenarios, the actual anomaly ratio in the dataset is not known. It is desired to have a model that works with a very conservative estimate of the anomaly ratio while still achieves reasonable accuracy.

Several interesting observations can be made by analyzing the results in Table \ref{tab:accuracy}. It is observed that considering feature space correlation clearly improves anomaly detection accuracy. More specifically, compared with the VAE-DN (without GCN to perform correlation pattern filtering), the proposed framework achieved almost 30-60\% improvement in terms of $F_1$ score on Thyroid, MoCap and UJIIndoorLoc datasets. This shows that considering 
feature space correlation makes a significant improvement on the detection accuracy when data features have strong internal correlation structure (e.g. interdependency of the physiological features under disease for Thyroid dataset, the correlated relative movement of the measuring points in the hand posture dataset MoCap and the underlying pattern of the relative position of the WiFi fingerprint signals in UJIIndoorLoc dataset.).

\subsubsection{Impact of the generated anomalous degree measures} It is found that the CSCAD consistently achieves higher $F_1$ score compared with the variant method CSCAD(no $\sigma$) that does not consider the latent standard deviation vector $\sigma_z$. This shows that considering only reconstruction loss may not fully reflect the anomalous behavior of a sample. Incorporating the internal uncertainty level information reflected in $\sigma_z$ is also important. This observation is also confirmed in Figure. \ref{dataset_vis}, which presents the visualization of detection results with respect to the L2-norms of reconstruction loss $d$ and latent standard deviation vector $\sigma_z$. It is observed that there does not exist simple boundary values to separate the normal and anomalous samples by solely inspecting $d$ or $\sigma_z$. However, by joint modeling both $d$ and $\sigma_z$, we can learn a more robust discriminative boundary to detect anomalies. 

\begin{table}[th]
	\centering
	\caption{Experiment results with limited anomaly label available for KDDCUP dataset.}
	\label{tab:anomaly}
	\scalebox{0.99}
	{
	\begin{tabular}{c|ccc}
		\toprule
		Labeled Rate    & Precision  & Recall    & $F_1$   \\ 
		\midrule
		0\%     & 0.881   & 0.996    & 0.934     \\ 
		0.50\%     & 0.883   & 0.990    & 0.934     \\ 
		0.75\%     & 0.883   & 0.997    & 0.936          \\ 
		1.00\%     & 0.881   & 1    & 0.937    \\  
		\bottomrule
	\end{tabular} 
	}
\end{table}

\subsubsection{Results under semi-supervised settings} CSCAD can be easily adapted to semi-supervised or supervised settings by using actual anomalous or normal samples in the positive and negative sample set during training the discriminative network. We conduct an experiment on the KDDCUP dataset to test the performance of our framework when limited amount of anomaly labels are known. Table \ref{tab:anomaly} presents the results of CSCAD(2.5\%) model when replacing 0\% to 1\% of the 2.5\% high anomalous degree negative samples with the actual anomalies. It is observed that with the introduction of higher amount of labeled anomalies, the $F_1$ score increases from $0.934$ to $0.937$. Even without labeled data, the framework trained under unsupervised setting already achieved reasonable accuracy ($0.934$), with only $0.003$ decrease in $F_1$ score compared with the case introducing 1\% of actual anomalies during training. This demonstrates the robustness of the proposed framework.

\begin{table}[th]
	\centering
	\caption{Experiment results of our framework and the baseline methods for Heterogeneity dataset}
	\label{tab:extension}
	\scalebox{0.99}{
	\begin{tabular}{c|ccc}
		\toprule
		\multirow{2}{*}{Methods} &  \multicolumn{3}{c}{Heterogeneity}                     \\ 
		& Precision & Recall & $F_1$             \\ 
		\midrule
		IF & 0.830        & \textbf{1}        & 0.907              \\ 
		SR-CNN &  0.834     & \textbf{1}        & 0.909              \\ 
		AE(LSTM)-IF     & 0.825        & 0.994    & 0.902                \\ 
		VAE(LSTM)-DN    & 0.953        & \textbf{1}        & 0.976           \\  
		\midrule
		CSCAD(LSTM)  & \textbf{0.980}        & \textbf{1}        & \textbf{0.990}               \\ 

		\bottomrule
	\end{tabular} 
	}
\end{table}
	
\subsubsection{Time-series extension} To demonstrate the extensibility of our framework, we also present the experiment results on the time-series dataset Heterogeneity in Table \ref{tab:extension}. 
We compare our framework (CSCAD(LSTM)) with a classic method (IF) and two deep learning-based approaches (AE(LSTM)-IF and VAE+DN(LSTM)). Our framework outperforms all the baseline methods in terms of precision, recall and $F_1$ score. Compared with IF and AE(LSTM)-IF, the proposed framework achieved 8-9\% improvement on $F_1$ score. Again, we observe the feature space correlation helps to improve detection accuracy, which accounts for 1.4\% increase on $F_1$ score compared with the VAE-DN (LSTM) model.
These results show the effectiveness and extensibility of our framework. 

\section{Related Work}
There is extensive literature related to anomaly detection. Our focus is mainly restricted to high-dimensional collective anomaly detection problem for multiple types of data with limited or no anomaly labels (see \cite{chandola2009anomaly,pimentel2014review,ahmed2016survey,zhang2020urban} for a wider scope survey).  In this scope, most recent works tend to utilize a reconstruction-based approach or evaluate the anomaly measure/score with a threshold (predetermined or learned) to solve the anomaly detection problem.

\subsection{Reconstruction-based Anomaly Detection}
The main assumption of the reconstruction-based approach is that anomalous samples have different patterns compared with normal samples, hence they are more difficult to be reconstructed. The anomalous degree of a data sample can be reflected by the loss or distance between the original and reconstructed data samples generated by some statistical or neural models. Classic methods include principal component analysis (PCA) with explicit linear projections, and the improved version, robust principal component analysis (RPCA) \cite{candes2011robust}, \cite{huber2004robust}, which makes PCA more robust by enforcing sparse structures. Inspired by RPCA and deep learning techniques, Zhou and Paffenroth \cite{zhou2017anomaly} introduce a robust deep autoencoder (RDA) model which split the input data into reconstructed part and noise to improve the robustness of standard deep autoencoders. Other methods \cite{deecke2018anomaly}, \cite{schlegl2017unsupervised}, \cite{li2018anomaly} detect anomalies using generative adversarial networks (GANs) \cite{goodfellow2014generative}. The idea is that anomalies differ from the distribution of normal samples, which makes it difficult to generate similar non-anomalous samples through GANs. Additionally, features in the multi-variational anomaly detection problem not always work independently. In these problems, each of the individual points in isolation appears as normal data instances while observed in a group exhibit unusual characteristics. A low-capacity model is unable to capture the complex patterns in the data, resulting in model-induced reconstruction deficiency. The deep anomaly detection techniques have been to learn complex hierarchical feature relations within high-dimensional raw input data \cite{lecun2015deep}. The number of layers used in the technique is driven by the input data dimension, a deeper network is needed to produce a better performance \cite{chalapathy2019deep}. Chalapathy el at. \cite{chalapathy2018group} focuses on mining irregular group distributions by utilizing Adversarial autoencoder (AAE) and variational autoencoder (VAE) for group anomaly detection. Araya el at. \cite{araya2016collective} train an autoencoder to capture the internal relationship among feature while recognizing normal consumption patterns in the training data. However, the weakness of these reconstruction-based approaches lies in the need of a reliable data reconstruction model

\subsection{Anomaly Score-based Anomaly Detection (predetermined)}
Most of the unsupervised studies have been done on performing anomaly detection based on some anomaly representation score calculated by conventional or deep learning models with a predetermined threshold to evaluate anomalies.
Zhang et al. \cite{zhang2018detecting} focus on detecting the traffic volume anomaly which mainly relies on the pairwise similarity matrix for each region and data source to define the anomaly score. We choose the deep learning architecture mainly depends on the nature of input data. Oh et al. \cite{oh2019sequential} proposed IRL-ADU which uses an RL-based method to solve the time series anomaly detection problem. IRL-ADU is an end-to-end framework for sequential anomaly detection using inverse reinforcement learning (IRL) which can learn the decision-making agent’s underlying function. 
Additionally, in most instances, data samples are considered as anomalous when they are located in a low density/probability region of the training data. In this case, we use the probability as the basis to calculate the anomaly score and judge the anomaly. Meanwhile, Input data depending on the number of features can be further classified into either low or high-dimensional data which is important for us to choose the different methods. Some traditional methods (eg. kernel density estimation \cite{parzen1962estimation} and Robust-KDE \cite{kim2012robust}) are known to be problematic in dealing with high-dimensional data due to the curse of dimensionality. To mitigate this problem, some studies \cite{candes2011robust} compresses the high-dimensional data into low-dimensional latent representations using deep autoencoders, and then applies a density-based model on the low-dimensional space to detect the anomaly. Some recent works combine these two steps and directly learning an anomaly score that perform density estimation. Zhai et al. \cite{zhai2016deep} utilize an energy-based autoencoder model to map each data sample to an energy score. Zong et al. \cite{zong2018deep} use a compression network combined with the Gaussian mixture model to estimate the density of each sample and further detect the anomaly. Other methods \cite{an2015variational}, \cite{xu2018unsupervised}, \cite{ikeda2018estimation} use a variant of VAE to learn the probability density of each sample as and identify the low-probability samples as anomalies.  These anomaly score-based methods are all relay on a predetermined threshold to do the final decision, it will limit the accuracy of the model when we know few about the anomaly labels.

\subsection{Anomaly Score-based Anomaly Detection (learned)}
A major drawback of many anomaly score-based methods is the need to specify some thresholds for the anomaly score or proportion of anomalies in the data to discriminate normal or anomalous samples, which involves additional assumptions on the data. To overcome this drawback, another anomaly score-based detection technics tries to learn the threshold by a data-driven approach. Ramakrishnan et al. \cite{ramakrishnan2019anomaly} utilize the use cross-validation on the test set to select the threshold in the experiments and choose the threshold that maximizes the recall at a minimum precision level in the production system. OmniAnomaly \cite{su2019robust} set the anomaly threshold offline following the principle of Extreme Value Theory (EVT) whose goal is to find the law of extreme values, and extreme values are usually placed at the tails of a probability distribution \cite{siffer2017anomaly}. But the above methods are all designed as unsupervised learning due to the lack of large-scale labeled anomaly data. As a result, they are difficult to leverage prior knowledge (e.g., a few anomaly labels) when some information is available as in the real-world anomaly detection applications. Pang et al. \cite{pang2019deep} designed a framework to leverage a few anomaly labels and a prior probability to enforce statistically significant deviations of the anomaly scores of anomalies from that of normal data objects in the upper tail. Ren et al. \cite{ren2019time} injecting anomaly points as anomaly labels into a collection of saliency maps that are not included in the evaluated data and training a discriminating network to classify the anomalies. Ren et al. proposed the SRCNN \cite{ren2019time} attempt to borrow the SR model from visual saliency detection domain to time-series anomaly detection and using the CNN as the discriminating network instead of a threshold to identify anomaly observations.

In this work, we combine the merits of the aforementioned approaches while overcomes their drawbacks. We exploit the internal correlation structure in data features and use a high-capacity model for data reconstruction. We further train a small discriminative network to perform detection using the data reconstruction loss and latent standard deviation. The proposed framework is highly adaptive, which can be used with limited or no anomaly labels, and easily generalizable to high-dimensional time-series data.

\section{Discussion and Conclusion}
We propose a new adaptive framework (CSCAD) for collective anomaly detection on high-dimensional data for a large complex system with limited or no anomaly labels. 
Unlike the state-of-the-art approaches so far, CSCAD jointly considers the correlation structure in the feature space and robust sample reconstruction, which leads to superior performance in high-dimensional collective anomaly detection tasks. 
To the best of our knowledge, this is the first attempt to use the graph convolutional network as a filter to capture the variation in correlation structure among features, which enables more expressive modeling of the normal patterns in data. 
In this framework, we propose a new EMI metric which is capable of evaluating the correlation of data with different types (continuous and categorical) and properties (static and time-series). A reconstruction network is developed to perform sample reconstruction and evaluates two natural anomalous degree measures of each sample: the reconstruction loss and the latent standard deviation. These two anomalous degree measures serve as inputs to a discriminating network to perform final anomaly detection, which is trained using high anomalous degree samples as positive samples, and low anomalous degree samples or samples with anomaly labels as negative samples. This scheme allows the discriminating network can be trained without a predetermined threshold and adaptable to semi- or supervised settings, which is a perfect fit for many real-world applications. 
Numerical results on five public datasets show that our framework consistently outperforms the baseline methods, which proves the effectiveness of CSCAD.

We believe the proposed CSCAD can build a bridge between the fields of correlation mining and the practical application of collective anomaly detection. The generated feature level anomalous degree measures can also be used in anomaly propagation tracing and anomaly source identification.
The proposed CSCAD framework is a scalable and highly flexible approach that can extend to other scenarios such as time-series and image anomaly detection tasks. 




%
%
%
%

\bibliographystyle{IEEETrans}
\bibliography{bare_adv}

%
%

\end{document}